\documentclass{INTERSPEECH2023}
\interspeechcameraready

\title{On Monotonic Aggregation for Open-domain QA}
\name{Sang-eun Han$^{1,2*}$\thanks{~~* First two authors equally contributed to this work.}, Yeonseok Jeong$^{1*}$, Seung-won Hwang$^{1,2\dagger}$\thanks{~~$\dagger$ Corresponding Author}, Kyungjae Lee$^3$}
\address{
  $^1$Seoul National University, Republic of Korea\\
  $^2$SNU-LG AI Research Center, Republic of Korea\\
  $^3$LG AI Research, Republic of Korea}
\email{\{gkstkddms00, jys3136, seungwonh\}@snu.ac.kr, kyungjae.lee@lgresearch.ai}

 \usepackage{times}
\usepackage{latexsym}

\usepackage[T1]{fontenc}

\usepackage[utf8]{inputenc}

\usepackage{microtype}
\usepackage{booktabs}
\usepackage{makecell}
\usepackage{graphicx}
\usepackage{mathtools}
\usepackage{enumitem}
\usepackage{float}
\usepackage{colortbl}
\usepackage[most]{tcolorbox}

\usepackage{pifont}
\usepackage{multirow}
\usepackage{amsmath}
\usepackage{amssymb}
\usepackage{subcaption}
\usepackage{hyperref}
\usepackage[toc,page]{appendix}

\hypersetup{colorlinks,allcolors=black}

\newcommand{\RNum}[1]{\uppercase\expandafter{\romannumeral #1\relax}}

\newcommand{\eg}{\textit{e}.\textit{g}.}

\usepackage{comment}
\usepackage{amsmath}
\usepackage{tabularx}
\usepackage{caption, booktabs}
\usepackage{algorithm}
\usepackage{algpseudocode}
\usepackage{amssymb}
\usepackage{setspace}
\usepackage{bbm}
\usepackage{arydshln}

\makeatletter
\newcommand\footnoteref[1]{\protected@xdef\@thefnmark{\ref{#1}}\@footnotemark}
\makeatother
\begin{document}

\maketitle





\begin{abstract}

Question answering (QA) is a critical task for speech-based retrieval from knowledge sources, by sifting only the answers without requiring to read supporting documents.
Specifically, open-domain QA aims to answer user questions on unrestricted knowledge sources.
Ideally, adding a source should not decrease the accuracy, but we find this property (denoted as ``monotonicity'') does not hold for current state-of-the-art methods. We identify the cause, and based on that we propose Judge-Specialist framework. Our framework consists of (1) specialist retrievers/readers to cover individual sources, and (2) judge, a dedicated language model to select the final answer. Our experiments show that our framework not only ensures monotonicity, but also outperforms state-of-the-art multi-source QA methods on Natural Questions. Additionally, we show that our models robustly preserve the monotonicity against noise from speech recognition. We publicly release our code and setting.


\end{abstract}

\section{Introduction}

Open domain
question answering (ODQA) aims to answer user questions 
on unrestricted topics and knowledge sources.
ODQA systems are especially useful when information is dispersed across sources, which makes it difficult for users to find relevant information quickly.
The goal of this paper is to improve ODQA techniques to enhance the efficiency and effectiveness of information retrieval through speech.

A major challenge in ODQA is 
to synergize multiple knowledge sources
such as unstructured text~\cite{izacard2021leveraging},
structured or semi-structured sources such as knowledge graph~\cite{agarwal-etal-2021-knowledge} and tables~\cite{nan2022fetaqa}
by leveraging their complementary strengths:
structured sources are highly precise but answer limited questions, while unstructured text answers a wider range of questions due to its widespread availability.
A similar challenge  is aggregated search \cite{lalmas2011aggregated} for multiple vertical search engines, \eg, image or table search.
Aggregated search consists of individual verticals,
and the aggregator which ranks the vertical results for user satisfaction.

To illustrate, consider a query from Natural Questions (NQ)~\cite{kwiatkowski2019natural}: \textit{``Who is the director of the film Avatar?''}.
This query can be answered by both text search and table search in the aggregated search, and the aggregator would decide to rank the result with a higher expected quality first.

\begin{figure}[t!]
     \centering
         \includegraphics[width=1.0\linewidth]{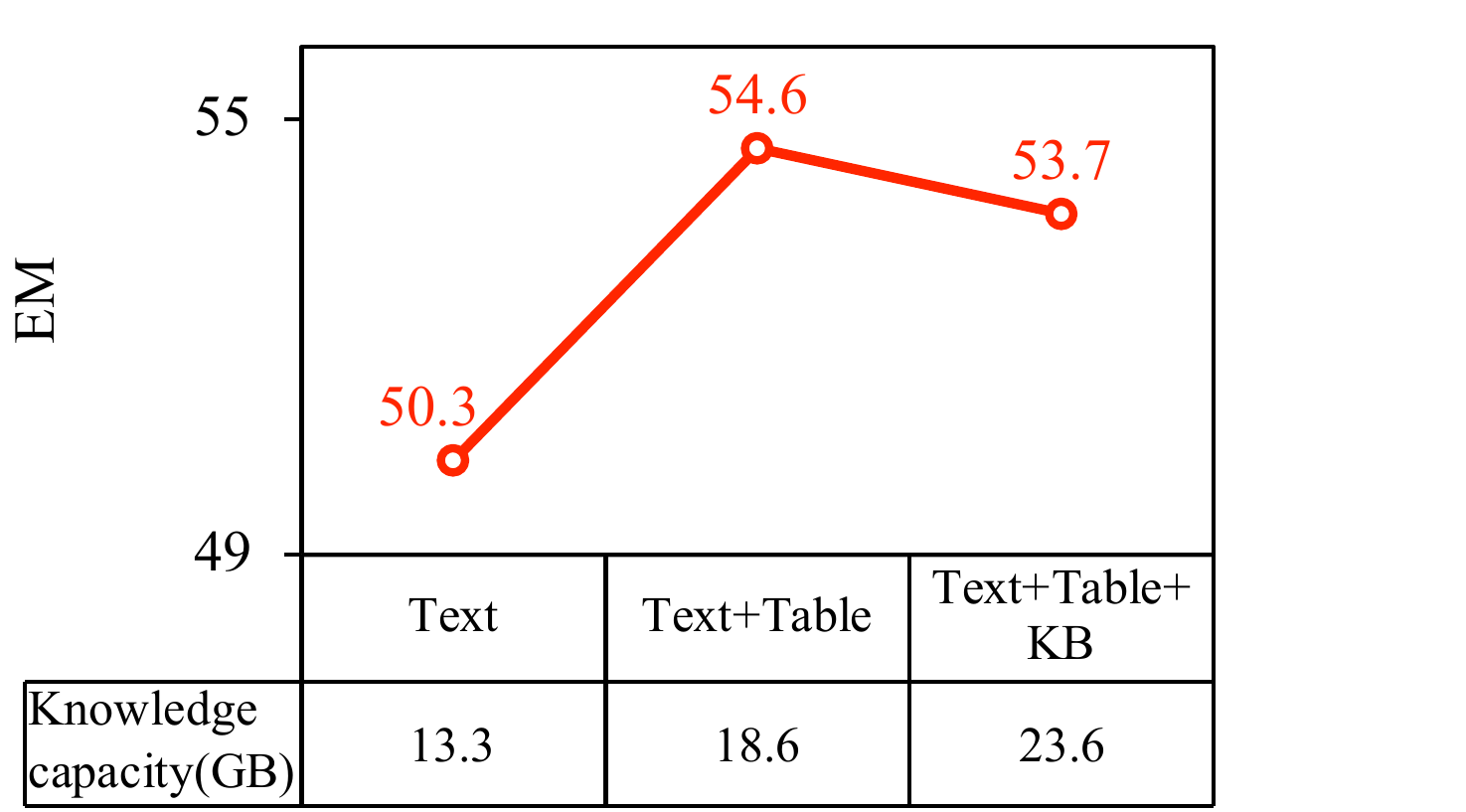}
     \hfill
\caption{Result in \cite{oguz2020unik}.
Added sources in prior work increase the required storage for knowledge, but EM accuracy decreases.
}
\label{fig:figure_1}
\end{figure}
Meanwhile, prior work on multi-source ODQA, UniK-QA \cite{oguz2020unik} takes a different approach.
UniK-QA unifies the representation of sources by linearizing graphs, tables, and lists into unstructured text,
such that a unified retriever-reader can identify an answer.
 For example, a table row containing the director of Avatar can be transformed into the text (\eg, \textit{``Director of Avatar is James Cameron''}), such that information from diverse sources can be jointly processed by retriever and reader.
 UDT-QA \cite{ma2022open} inherits this approach, and unifies the sources with a pretrained language model (PLM).
This line of work has shown a significant improvement over single-source QA system, demonstrating the benefit of incorporating multiple knowledge sources in ODQA.
However, we argue that previous work fails to exploit the increased capacity and diversity of knowledge sources.
To motivate, Figure \ref{fig:figure_1}
shows that the performance (measured as Exact Match, or EM)
does not monotonically improve with increased sources, as the addition of knowledge base (KB) results in performance degradation.

We refer to this phenomenon as ``non-monotonicity'',
and pursue the multi-source ODQA with ``monotonicity''
where the addition of knowledge sources does not damage the performance.
Our analysis in Section \ref{sec:causeofnm} shows that the unified retriever fails to cover multiple sources effectively, thereby causing non-monotonicity.
To address this issue,
we revisit and explore the framework of aggregated search in ODQA.
We keep source-specific retriever-readers as vertical `specialists', and propose `judge' to aggregate results for non-decreasing performance.
We denote this model as JS-HR, for Judge-Specialist model for High-Resource scenarios.
For low-resource scenarios where JS-HR is not affordable, we employ
multitask learning (MTL) to keep resource requirement near constant over the increase of sources.
We denote this model as JS-LR in Figure 2,
from which we confirm that monotonicity holds for both JS-HR and JS-LR:
as the sources are added from T (text), TT (text and table), and TTK (text, table and knowledge base), the performance increases accordingly.

Our empirical results also suggest that our proposed approach ensures monotonicity and outperforms the prior method in
NQ-speech dataset as well, representing realistic noises from speech recognition.
Our code and results are available at \texttt{github.com/YeonseokJeong/Judge-Specialist}.



\begin{figure}[t!]
     \centering
         \includegraphics[width=0.9\linewidth]{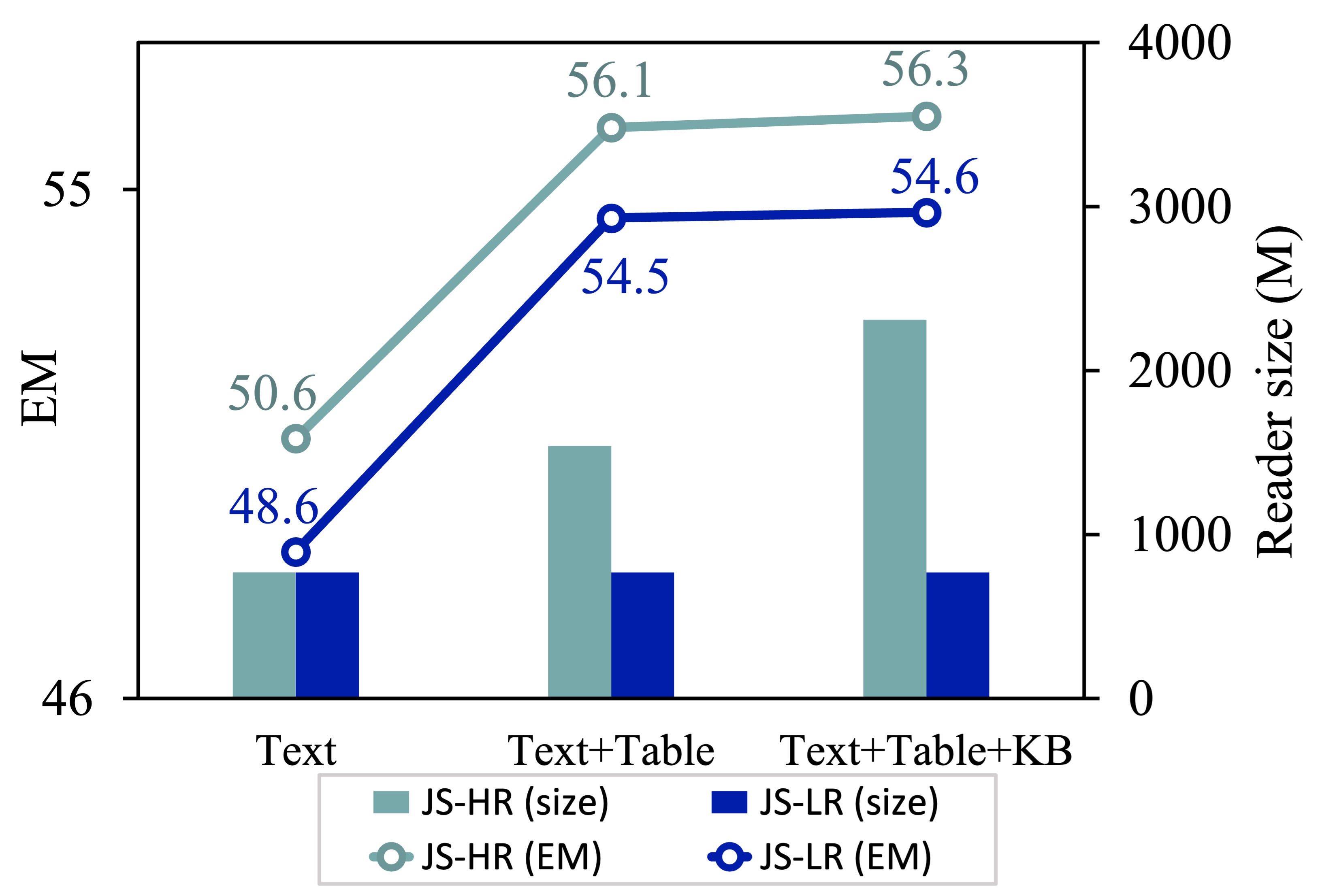}
     \hfill
\caption{Performance (line) and total parameter size for readers (bar) of our methods in NQ. }
\label{fig:figure_2}
\end{figure}

\section{Background \& Related Work}

This section overviews preliminary knowledge and existing work on
ODQA, the task of answering general domain questions from 
one or more knowledge sources.

\subsection{\textbf{ODQA with multiple knowledge sources}}
UniK-QA \cite{oguz2020unik} proposed to unify knowledge sources into textual format, such that simple transformer-based language models (LM) can support all sources.
In UniK-QA, structured data such as table and KB are unified with unstructured text by \textit{linearization};
cell values on the same row are concatenated into text by predefined rules. 
The unified knowledge is then used in DPR-FiD pipeline (explained in \ref{sec:retread}), where the retriever collects passages from multiple sources.

UDT-QA \cite{ma2022open} unifies knowledge sources into unstructured text as well, but with LM-based method known as \textit{verbalization}.
In verbalization \cite{ribeiro2020investigating}, PLM generates the unstructured text which describes the content of given structured data.\footnote{In UDT-QA, T5-large is used to verbalize structured data into text. To keep the faithfulness of generation against hallucination / missing information, training samples are filtered with ROUGE-1 score.}
The unified knowledge is then utilized in DPR-UnitedQA \cite{cheng2021unitedqa} pipeline.
We adopt the same verbalization approach for our work,
since the authors report that the verbalized knowledge is favored by readers over the linearized one.
UDT-QA outperforms UniK-QA on NQ, but the issue of non-monotonicity persists.

\subsection{\textbf{Retrievers and Readers}}
\label{sec:retread}
Retriever-reader architecture is conventionally adopted approach for ODQA \cite{izacard2021leveraging,oguz2020unik,cheng2021unitedqa}.
We build our model on DPR-FiD architecture, since it is widely adopted by state-of-the-art multi-source ODQA.
FiD models are trained from T5-large huggingface checkpoint.

DPR, or dense passage retriever, is widely used for the document retrieval, such as in \cite{izacard2021leveraging,oguz2020unik,ma2022open}.
DPR demonstrated significant improvement over the previous methods such as TF-IDF or BM25 \cite{robertson2009probabilistic}.
DPR retriever has the biencoder structure, which consists of a document encoder and a query encoder.
The encoders are trained such that the embeddings of positive documents have greater dot product with the question embedding.

FiD, or fusion-in-decoder, is a commonly used reader component in ODQA, due to its ability to effectively fuse information from multiple passages \cite{oguz2020unik,lewis2021paq}.
Implemented on encoder-decoder model such as T5 \cite{raffel2020exploring} or BART \cite{lewis2019bart}, FiD encodes retrieved passages independently, then the decoder computes full attention on the concatenated representation of passages.

\textbf{Our distinction} is to
separate retrievers and readers to obtain answer candidates from each specific source,
instead of relying on unified retriever-reader pipeline.
The judge then filters out incorrect candidates to preserve the monotonicity.
Our approach is comparable to Watson \cite{ferrucci2012watson}, where multiple expert systems produce candidates
then aggregate them to determine the final answer.
However, while Watson relies on multiple different low-level algorithms such as relation extraction,
we leverage the ability of transformer-based LM (T5-large) to evaluate the candidates with unified knowledge.

\section{Proposed Method}

This section overviews JS-HR (Judge-Specialist for High Resource) optimizing for monotonicity (Section 3.1) and JS-LR (Judge-Specialist for Low Resource) for a higher scalability (Section 3.2).

\subsection{Goal 1: Monotonicity}
To preserve monotonicity, QA process
 is broken down into two phases: \textbf{V-step} to collect enough answer candidates from vertical specialists,
 and \textbf{A-step} to aggregate the candidates and determine the final answer with the judge model.

 \begin{figure*}[!t]
 	\centering
 	\includegraphics[width=0.89\linewidth]{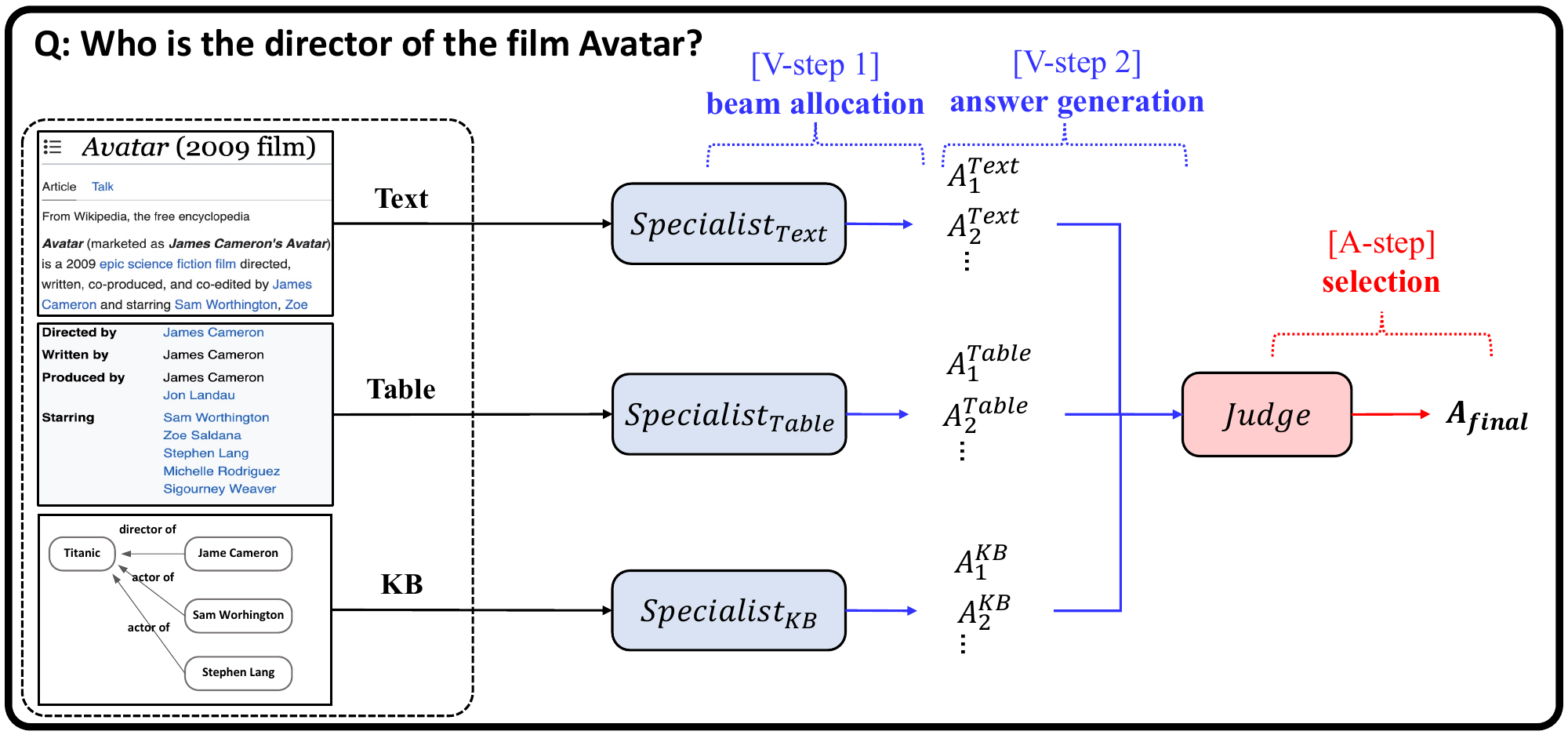}
 	\caption{Overview of our method (JS-HR). Multiple \textit{specialists} generate candidates, and one \textit{judge} selects the final answer. For judge model, we use fine-tuned T5-large.}
 	\label{method_overview}
\end{figure*}




\subsubsection{V-step: Candidates from Specialists}
\label{sec:v-step}

The V-step aims to collect candidates from each vertical specialist, thereby improving the recall of the system.
We use beam search to generate candidates,
and invest a larger beam size to high-performing specialists,
such that a more useful source can provide more candidates.
Specifically, beam size is set to be proportional to the
dev set performance (EM) of the specialist reader representing a source.

We first determine the proportion of beam size for each source, then allocate the beam size according to the manually chosen total budget
(total number of candidates).
Formally, the proportion $p_k$ for $k$-th source among $m$ sources is calculated as below:

\begin{equation}
     p_k = \frac{EM_k}{\sum_{j=1}^m{EM_j}} \\
\end{equation}
where $EM_x$ is the dev set EM of the specialist reader representing $x$-th source.
Beam size $b_k$ is then calculated with regard to the given total budget $B$.
\begin{equation}
        b_k = \lceil p_k  \cdot B \rceil
    \label{static}
\end{equation}
We round  $b_k$ to the nearest positive integer.

\subsubsection{A-step: Selection with the Judge}

In A-step, we aggregate candidates $ <A_{1}, A_{2}, ..., A_{n}>$ from verticals and select the final answer.
We frame the selection process as a problem of calibration,
quantifying the trustworthiness of candidates with probability $P$ from LM.
Formally, candidate answer $A$ which consists of token $a_{1}, a_{2}, ..., a_{n}$ can be
evaluated by LM probability $P$.
\begin{equation}
    P(A | Q ,C) =
    \{\prod_{t=1}^{n} P(a_t|a_{<t}, Q, C)  \}^{\frac{1}{n}} 
    \label{eq2}
\end{equation}
where $Q$ is question and $C$ is the  context, which can be obtained from both
existing unified retriever\footnote{We use the same unified retriever across all multi-source settings (TT, TTK). Specifically, we use unified retriever trained / inferred on TT data for our main experiments.} and our proposed specialist retriever.

The two types of retrievers have complementary strength.
If the answer is unanimously supported by all sources,
the unified retriever can provide high-quality context.
To illustrate, in Figure \ref{method_overview}, specialist retriever for text would bring relevant passage \textit{``Avatar is a SF movie directed by James Cameron''},
and similarly specialists for table/KB would retrieve verbalized evidences separately.
Alternatively, unified retriever would integrate evidences from diverse sources: ``\textit{Avatar is a SF movie directed by James Cameron.} (text) $\cdots$ \textit{Director of Avatar is James Cameron.} (verbalized table)''

Meanwhile, we can imagine another extreme where only one of the sources supports the answer.
In such case, passages from non-supporting sources in the unified retriever would only work as a noise.
For such cases, using context from specific source has the effect of denoising unrelated information.

We evaluate candidates with both types of context, and combine two LM probabilities for the final selection.
We provide the context from unified retriever
to the judge, and calculate the probability $P_{J}$ for $ A_{1}, A_{2}, ..., A_{n}$ as Equation \ref{eq2}.
Similarly, context from specialist retriever is provided to the specialist reader, to calculate the probability $P_{S}$ for each candidate.
We then select the candidate with the highest average value of $P_J$ and $P_S$ from the candidate set.
While there are multiple ways to combine the probabilities, we empirically found that a simple average gives strong results.

\subsection{Goal 2: Graceful Scaling}

We discuss how to implement the Judge-Specialist framework, for both high- and low-resource scenarios.

\textbf{JS-HR} (Judge-Specialist for High-Resource):
By default, we utilize the FiD \cite{izacard2021leveraging} models for both the specialist readers and the judge, which are trained on the passages of respective sources.
Additionally, we assume that each specialist operates independently and is trained in isolation.
This setting is referred to as JS-HR, which is the original version of our Judge-Specialist framework.

\textbf{JS-LR} (Judge-Specialist for Low-Resource):
In an effort to reduce the size of specialists,
we can consider diverse MTL techniques.
We stress that JS-LR is orthogonal to any such technique,
and implement JS-LR with a widely adopted baseline:
we simultaneously feed single-source contexts to the model during training, and backpropagate the sum of losses from each source.
This integrated model requires 3 times fewer parameters than an original design while still serving as a specialist for any source.


\section{Experimental Results}

\subsection{Main Results}
\begin{table}
\centering
\begin{tabular}{lcccccc} 
\noalign{\hrule height 1pt} 

\multicolumn{1}{c}{\multirow{2}{*}{\textbf{Model}}}  & \textbf{Text} & \textbf{Text+Table}   & \textbf{TT+KB}         \\ 
\multicolumn{1}{l}{} & (T) & (TT) & (TTK)\\
\hline
\textit{Prior methods} & \\
~~ UniK-QA (multi-data)              & 50.3 & 54.6 & 53.7       \\
~~ UniK-QA (per-data)              & 49.0 & 54.1 & 54.0       \\
~~ UDT-QA               & \textbf{51.8} & 55.2 & 55.1         \\
~~ Baseline           & 50.9 & 53.7 & 53.0      \\
\hline
\textit{Ours} & \\
~~ JS-HR           & 50.6 & \textbf{56.1} & \textbf{56.3}      \\
~~ JS-LR              & 48.7 & 54.5 & 54.6          \\
\noalign{\hrule height 1pt} 
\end{tabular}
\caption{Experimental result on the test sets of NQ using EM scores (with budget B=9)}
\label{main_result}
\end{table}


\label{sec:main_result}
Table \ref{main_result} shows the result of our models
on NQ,\footnote{
We use open-domain split of NQ,
which contains 79K examples for train set, 8.7K for dev set, and 3.6K for test set.
Following \cite{ma2022open}, we use the knowledge-answerable data augmented train set for TT / TTK reader, which consists of 80K examples.
We report single-run result (both training and evaluation) on this dataset.
} compared to existing methods.
We also compare ours to the baseline model which consists of UDT-QA retriever and FiD reader
(referred to as `Baseline' in Table \ref{main_result}),
to better observe the effect of the separated retrievers.
The result shows that (1) JS-HR outperforms existing methods (including baseline) in multi-source settings and
(2) JS-HR and JS-LR both preserve the monotonicity,
whereas other methods fail to benefit from KB.
\footnote{
We use 8 A6000 GPUs for the beam search, and 8 RTX 3090 GPUs for other experiments.
Each reader is trained with learning rate of 5e-5 and 10 epoch for 6 days, and the best checkpoint is selected by the corresponding dev EM.
We apply early stopping when the dev EM does not increase within two consecutive epochs.
}

\begin{table}
\centering
\begin{tabular}{lll} 
\noalign{\hrule height 1pt} 
\textbf{Method}  & \textbf{JS-HR} & \textbf{JS-LR} \\
\hline
$P_S$         & 51.0  & 50.9  \\
$P_J$         & 55.3  & 54.5  \\
$P_{JS}$ \ (ours) & \textbf{56.3}  & \textbf{54.6} \\
\noalign{\hrule height 1pt} 
\end{tabular}
\caption{Ablation study on the selection in NQ using EM score (TTK).}
\label{tab:RQ1}
\end{table}

\begin{figure}[t!]
     \centering
         \includegraphics[width=0.9\linewidth]{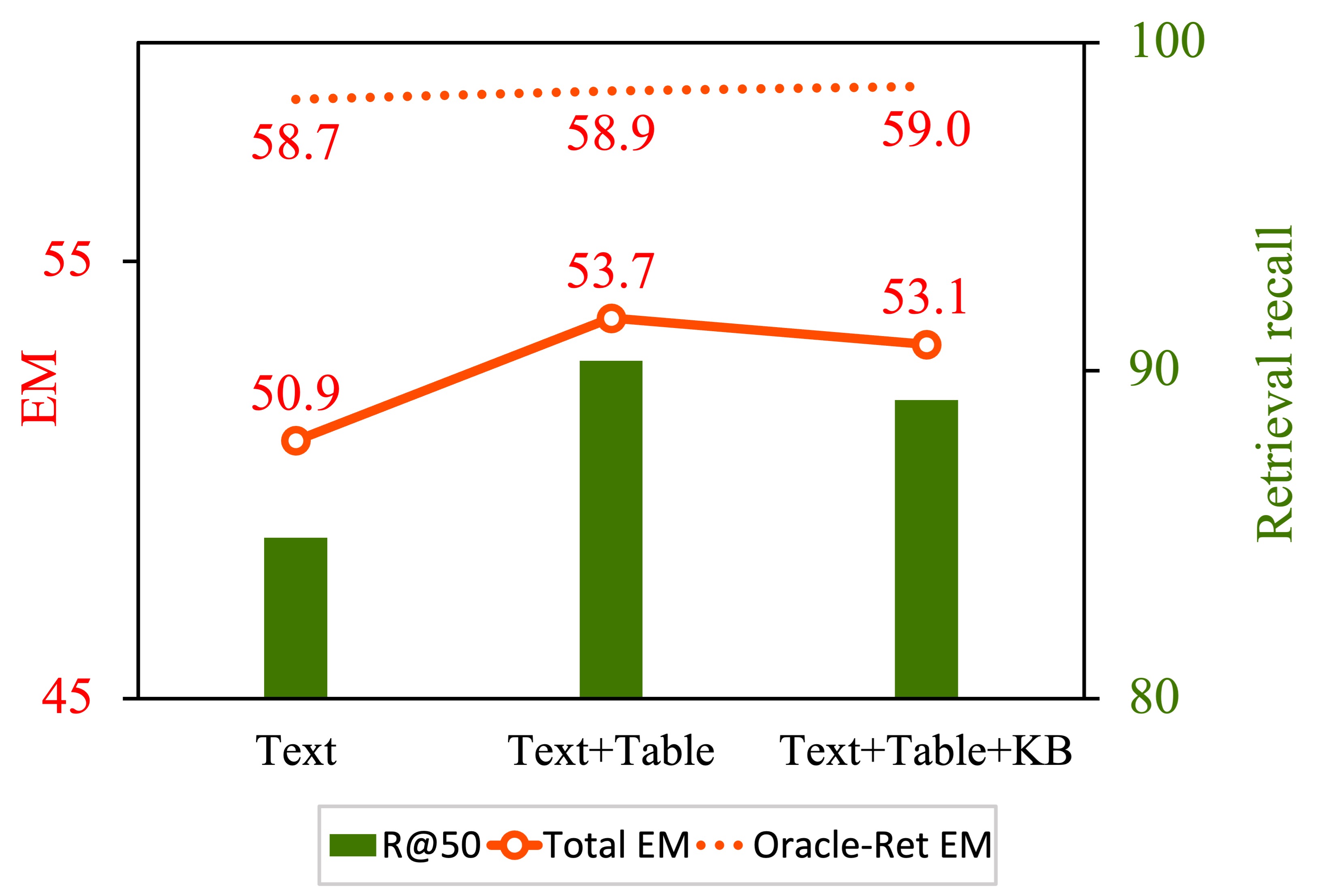}
     \hfill
\caption{Analysis of non-monotonicity of a baseline in NQ (TTK). }
\label{fig:figure_nm}
\end{figure}

\subsection{Cause of non-monotonicity}
\label{sec:causeofnm}
Figure \ref{fig:figure_nm} shows our analysis on the unified retriever-reader pair, using the same retriever setting as \cite{ma2022open}
(baseline method discussed in Section \ref{sec:main_result}).
Our analysis reveals that the non-monotonicity results from the failure of the unified retriever.

We observe that the monotonicity holds in NQ subset, where the retriever successfully obtains the oracle context that contains the ground truth. (shown by the `Oracle-Ret EM', dotted line in the figure)
However, without such condition, the retrieval performance (`R@50'\footnote{R@50 indicates the retrieval recall within the top-50 passages.}, bar graph) decreases as KB is added, leading to the non-monotonicity in overall QA performance (`Total EM', solid line). 
This result suggests that the unified retriever has a limited generalizability towards multiple sources, which motivates us to use the specialist retrievers.



\subsection{Effects of Judge}
 Table \ref{tab:RQ1} shows the ablation study on the selection.
 We observe dramatic performance drops in all settings when $P_{J}$ is removed from the selection criterion, 
 showing the critical role of the judge.
Meanwhile, the performance drop is limited when $P_{S}$ is removed.

\begin{table}[h]
\centering
\begin{tabular}{cccc} 
\noalign{\hrule height 1pt} 
\multicolumn{1}{l}{\textbf{Model}}  & \textbf{T} & \textbf{TT}   & \textbf{TTK}           \\ 
\hline
Baseline               & 29.8 & 32.2 & 31.5         \\
JS-LR              & 29.3 & 33.1 & 33.7          \\
JS-HR           & \textbf{30.4} & \textbf{33.8} & \textbf{34.0}      \\
\noalign{\hrule height 1pt} 
\end{tabular}
\caption{Experimental result on the test sets of NQ-speech. (with budget B=9)}
\label{tab:asr_result}
\end{table}

\subsection{Robustness to Speech Noises}
To deploy QA models with speech interface,
it is important to consider whether the models can generalize to inputs with real-life noises.
To examine the robustness to such noises,
we tested our models in NQ-speech where questions contain
errors from automatic speech recognition (ASR) \cite{sidiropoulos2022impact}.\footnote{ASR error is obtained by synthetic TTS system (Google TTS) and transcription system (wav2vec 2.0).}
Table \ref{tab:asr_result} shows that our models preserve monotonicity with speech noises, and generalize better than the baseline method discussed in Section \ref{sec:main_result}.

Additionally, we made a consistent observation in another realistic noisy dataset, NQ-typo,
where the questions have random typo errors \cite{sidiropoulos2022analysing}.
Our models perform closely to clean NQ, 
achieving $39.1$, $44.0$, $44.2$ in EM scores for JS-HR in T, TT, TTK respectively.
These results suggest
that our models show a higher generalizability in both settings, while their relative performance may depend on the noise type -- and ASR noise is more challenging.
A promising future direction is thus to close such a gap using speech-targeted debiasing \cite{sidiropoulos2022impact}.
\section{Conclusion}

In this paper, we proposed a simple yet effective method to utilize the complementarity in heterogeneous sources,
and thereby achieve the monotonicity,
unlike the current state-of-the-art methods.
Our proposed approach
keeps readers and retrievers as specialists to each source and adds a dedicated LM (the `judge') to aggregate candidates for the final answer. 
We validated the effectiveness and monotonicity in both clean and speech settings.







\section{Limitation}
\label{sec:limitation}




The focus of this work is to show the potential of Judge-Specialist framework, not to optimize the judge/specialist model.
Hence we did not further explore for the optimal practice in this paper,
but any such effort can be orthogonally combined with our framework.

\section{Acknowledgement}
\label{sec:acknowledgement}

This work was supported by LG AI Research.
This work was partly supported by Institute of Information \& communications Technology Planning
Evaluation (IITP) grant funded by the Korea government(MSIT) [NO.2021-0-01343, Artificial Intelligence Graduate School Program (Seoul National University)].

\bibliographystyle{IEEEtran}
\bibliography{mybib}

\end{document}